\documentclass[letterpaper, 10 pt, conference]{ieeeconf}  

\makeatletter
\let\NAT@parse\undefined
\makeatother

\IEEEoverridecommandlockouts                              

\usepackage[numbers, sort]{natbib}

\usepackage{balance}
\usepackage{hyperref}

\usepackage{ifthen,version}
\newboolean{include-notes}
\setboolean{include-notes}{false}
\newcommand{\mpnote}[1]{\ifthenelse{\boolean{include-notes}}%
{\textcolor{orange}{\textbf{MP: #1}}}{}}
\newcommand{\egnote}[1]{\ifthenelse{\boolean{include-notes}}%
{\textcol[1, 2, 3], tor{green}{\textbf{EG: #1}}}{}}
\newcommand{\bbnote}[1]{\ifthenelse{\boolean{include-notes}}%
{\textcolor{purple}{\textbf{BB: #1}}}{}}
\newcommand{\todo}[1]{\ifthenelse{\boolean{include-notes}}%
{\textcolor{red}{\textbf{TODO: #1}}}{}}
\newboolean{anonymized}
\setboolean{anonymized}{false}

\newcommand\figref{Fig.~\ref}
\newcommand\secref{Sec.~\ref}

\usepackage[nolist]{acronym}
\begin{acronym}
\acro{EIG}{expected information gain}
\acro{SLAM}{simultaneous localization and mapping}
\acro{SfM}{structure from motion}
\acro{AE}{active exploration}
\acro{MLE}{maximum-likelihood estimate}
\acro{bCH}{bidirectional Chamfer distance}
\acro{QP}{quadratic program}
\end{acronym}

\usepackage{amsmath,amssymb,amsfonts}
\usepackage{cancel}
\usepackage{nicefrac}

\newcommand{\field}[1]{\mathbb{#1}}

\newcommand{\E}{\mathbb{E}} 

\newcommand{\R}{\field{R}} 

\newcommand{\normsq}[1]{\left\|#1\right\|^2}
\newcommand{\twonorm}[1]{\|#1\|_2}

\newcommand{\twonormsq}[1]{\|#1\|_2^2}

\newcommand{\Loss}{\mathcal{L}}
\newcommand{\Data}{\mathcal{D}}

\usepackage{graphicx}
\graphicspath{ {images/} }
\usepackage{color} 

\usepackage{listings}
\definecolor{codegreen}{rgb}{0,0.6,0}
\definecolor{codegray}{rgb}{0.5,0.5,0.5}
\definecolor{codepurple}{rgb}{0.58,0,0.82}
\definecolor{backcolor}{rgb}{0.95,0.95,0.92}
\lstdefinestyle{python}{
    backgroundcolor=\color{backcolor},   
    commentstyle=\color{codegreen},
    keywordstyle=\color{magenta},
    numberstyle=\tiny\color{codegray},
    stringstyle=\color{codepurple},
    basicstyle=\footnotesize,
    breakatwhitespace=false,         
    breaklines=true,                 
    captionpos=b,                    
    keepspaces=true,                 
    numbers=left,                    
    numbersep=5pt,                  
    showspaces=false,                
    showstringspaces=false,
    showtabs=false,                  
    tabsize=2,
    extendedchars=true,
    literate={σ}{{$\sigma$}}1,
}

\title{\LARGE \bf
Active Tactile Exploration for Rigid Body Pose and Shape Estimation
}

\ifthenelse{\boolean{anonymized}}{
\author{Author Names Omitted for Anonymous Review$^{1}$
\thanks{$^{1}$Author affiliations and emails omitted for anonymous review}%
}
}{
\author{Ethan K. Gordon, Bruke Baraki, Hien Bui, Michael Posa$^{1}$
\thanks{$^{1}$All authors are with the General Robotics, Automation, Sensing, and Perception (GRASP) Laboratory,
        University of Pennsylvania, Philadelphia, PA, USA 19104,
        {\tt\small \{ethankg, bbaraki, xuanhien, posa\}@seas.upenn.edu}}%
}
}

\begin{document}

\maketitle
\thispagestyle{empty}
\pagestyle{empty}

\begin{abstract}

General robot manipulation requires the handling of previously unseen objects. Learning a physically accurate model at test time can provide significant benefits in data efficiency, predictability, and reuse between tasks. Tactile sensing can compliment vision with its robustness to occlusion, but its temporal sparsity necessitates careful online exploration to maintain data efficiency. Direct contact can also cause an unrestrained object to move, requiring both shape \emph{and} location estimation. In this work, we propose a learning and exploration framework that uses only tactile data to simultaneously determine the shape and location of rigid objects with minimal robot motion. We build on recent advances in contact-rich system identification to formulate a loss function that penalizes physical constraint violation without introducing the numerical stiffness inherent in rigid-body contact. Optimizing this loss, we can learn cuboid and convex polyhedral geometries with less than 10s of randomly collected data after first contact. Our exploration scheme seeks to maximize Expected Information Gain and results in significantly faster learning in both simulated and real-robot experiments.
\ifthenelse{\boolean{anonymized}}{
Website removed for anonymous review.
}{
More information can be found at: \url{https://dairlab.github.io/activetactile}.
}
\end{abstract}


\section{Introduction}

Robot manipulators in the wild will inevitably come across previously unseen objects and environments. They can benefit from building object models from the limited data available at test time. Such models can be reused across different tasks and systems, and they can make learned manipulation policies more interpretable.
In the building of these models, tactile sensing can compliment visual modalities like RGB and depth. Tactile data is robust to darkness, reflections, and occlusions from clutter, the environment, and the object itself. With the advent of inexpensive sensors \cite{lambeta_digit_2020, johnson_microgeometry_2011, do_densetact_2023}, touch is becoming a more common component of robot manipulation systems \cite{weinberg_survey_2024}. 

Two challenges inherent to tactile sensing are sparsity and disturbance. By definition, data is only collected about the part of the object in direct physical contact with the sensor, necessitating multiple movements to get a complete picture. Sparsity can be addressed by \emph{active exploration}, where actions are taken to efficiently maximize some information metric, often captured by maintaining a belief distribution over possible object parameterizations \cite{hejna_robot_2025, schneider_active_2022, yi_active_2016, driess_active_2017, zhang_active_2017, shahidzadeh_actexplore_2024}. However, reasoning about information gain is complicated by disturbance: every contact has the potential to move the object unless it is heavy or bolted down. Propagating belief through motion is tricky, informing our first key insight: \ac{EIG}, by foregoing an explicit belief distribution, is a particularly useful metric for active tactile exploration.

Disturbance affects even the model-building itself, as it necessitates learning the object's pose trajectory in addition to its geometry. That trajectory needs to adhere to the physical constraints of rigid-body motion and contact, which can lead to difficult loss landscapes with both nearly flat and nearly discontinuous regions. Related work in geometry-learning with a known trajectory \cite{pfrommer_contactnets_2020, bianchini_generalization_2022} has shown how a \emph{violation-implicit} loss leads to a landscape more amenable to gradient-based optimization. Our second key insight is that this loss works just as well with the trajectory itself as a decision variable.

\begin{figure}[t]
    \centering
    \includegraphics[width=\linewidth]{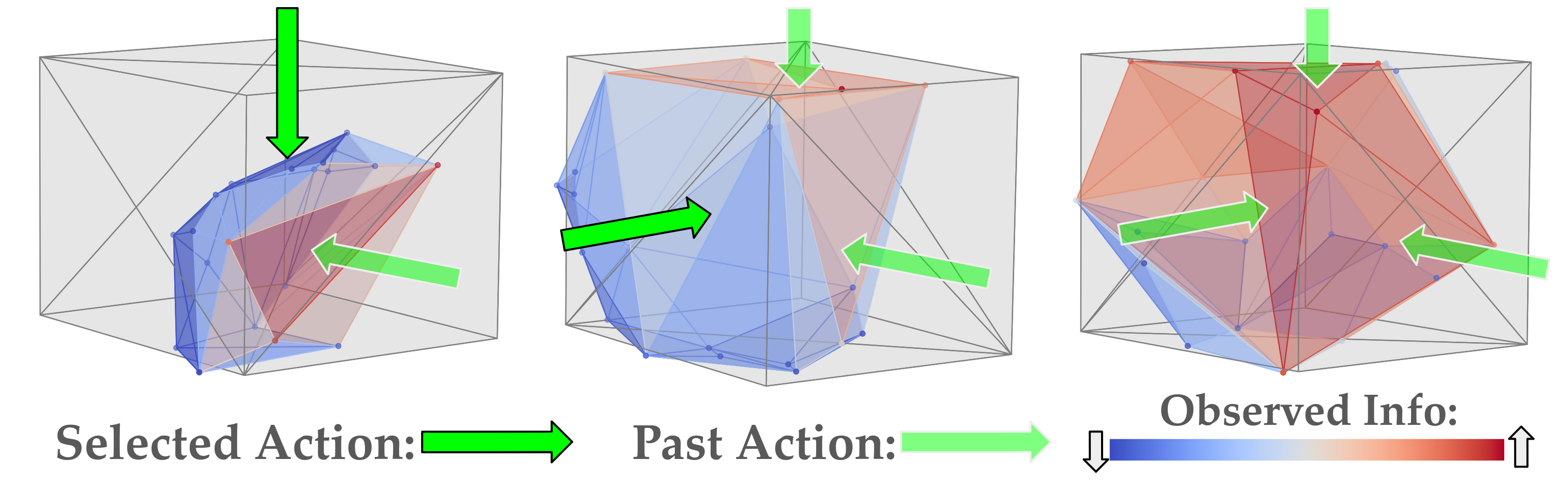}
    \caption{Only tactile data is used to find the pose and geometry of an arbitrary convex object (e.g., parameterized as a 20-vertex polytope as shown here). Actions are chosen to maximize \emph{Expected Information Gain}, related to the info we expect to gain from contact divided by the info previously observed. As parts of the ground-truth object (e.g. a cuboid, shown in gray) are contacted, the corresponding areas of the learned estimate will be assigned a high observed info (blue $\rightarrow$ red).}
    \label{fig:intro}
    \vspace{-0.1cm}
\end{figure}

In summary, this work presents two contributions.
\begin{enumerate}
    \item A \emph{violation-implicit loss} that enables the simultaneous leaning of rigid-body pose and geometry from purely tactile data.
    \item An active exploration scheme that seeks to maximize \emph{\ac{EIG}} to enable more efficient learning with fewer robot motions, as highlighted in \figref{fig:intro}.
\end{enumerate}
The combination of these contributions yields a scheme that can simultaneously learn the pose and cuboid or convex polytope geometric approximations of arbitrary unknown convex rigid bodies with under 10s of purely tactile data.

\begin{figure*}[t]
    \centering
    \includegraphics[width=\linewidth]{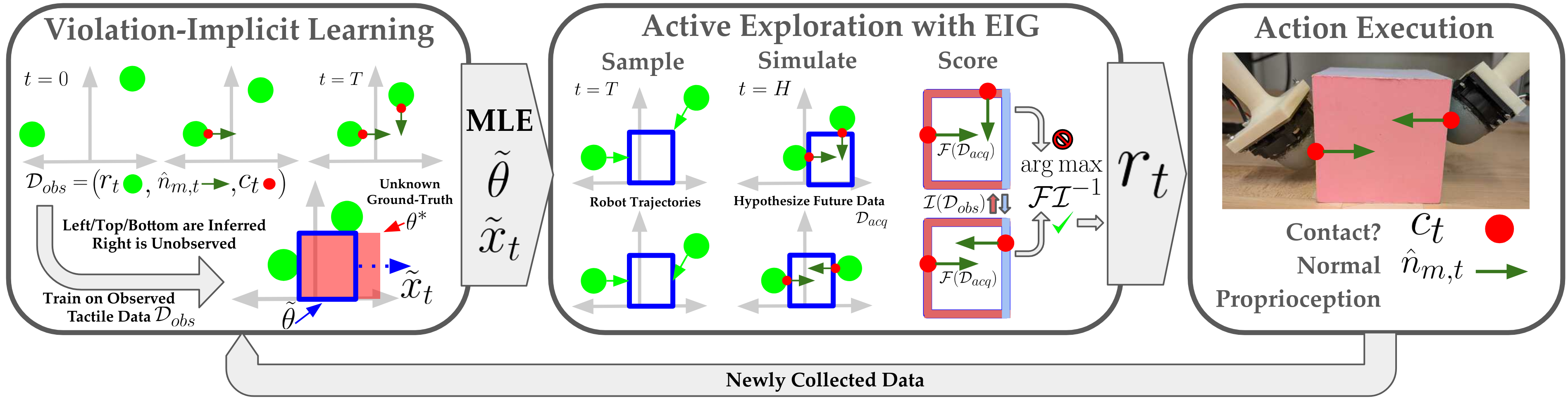}
    \caption{Block diagram of the interaction protocol. An estimate of object pose and geometry emerges from tactile and proprioceptive data (top left) by minimizing a violation-implicit loss. In this example with a cuboid parameterization, the top, left, and bottom faces can be directly inferred from data while the right face is unobserved. The expected information gain (EIG) of the next action is optimized using a Gaussian cross-entropy method: robot trajectories are sampled, simulated forward to time horizon $H$, and scored. Pictured here, $\Data_{obs}$ includes contact with left and top, leading to high observed info $\mathcal{I}$ (red), so the upper sample contacting these same sides yields low EIG. The lower sample contacts the right side with low $\mathcal{I}$ (blue), maximizing EIG.}
    \label{fig:block_diagram}
    \vspace{-0.1cm}
\end{figure*}

\section{Related Work}
\subsection{Model Building and System Identification}
Visual modeling of both environment and object geometry is a mature field with the well-studied paradigms of \ac{SLAM} and \ac{SfM}\cite{ozyesil_survey_2017}. Continuous object representations include radiance fields \cite{jiang_fisherrf_2024} and Gaussian splatting \cite{jiang_ag-slam_2024}. Recent work learns the geometry \cite{wen_bundlesdf_2023} and pose \cite{wen_foundationpose_2024} of dynamic objects in video, even in contact-rich environments \cite{bianchini_vysics_2025}. For assessing the accuracy of our tactile-only system, we compare against ground truth estimates from FoundationPose \cite{wen_foundationpose_2024}.

Even without active exploration, tactile sensing has been used in recent work to refine models of both the robot and its environment. Non-optical contact sensors can enable more accurate proprioception in the presence of a known environment \cite{koval_pose_2015}. Optical tactile sensors create estimates of local surface geometry that can be used to localize the robot with respect to a known object \cite{suresh_midastouch_2023} or vice versa \cite{sodhi_patchgraph_2022}.

\subsection{Active Exploration in Robotics}
\label{sec:related_ae}
Active exploration has been studied in robotics for decades \cite{bajcsy_active_1988} and involves either the selection of data-gathering actions to maximize information or the curation of data based on information potential. Generally, it includes: (1) some current state of knowledge, (2) an estimate of the information collected by a given action, and (3) an optimization metric quantifying how useful the latter is given the former.

In Bayesian state estimation, the current state can be represented by a belief distribution. Mutual information (MI) \cite{hejna_robot_2025} as a metric captures the information gained about this belief distribution by observing some other (e.g. measurement likelihood) distribution, but in dynamic environments this can be intractable to compute exactly \cite{schneider_active_2022}. Explicit MI calculation can be avoided by using uncertainty instead, e.g., probing areas where the belief distribution has a high variance \cite{yi_active_2016, driess_active_2017}. One could also use a heuristic to avoid actions similar to those previously taken \cite{shahidzadeh_actexplore_2024} or a policy trained to take informative actions \cite{zhang_active_2017}. For dynamic objects, propagating belief though a dynamics model that is itself uncertain is tricky: the result is inherently multimodal, and the number of required parameters scales exponentially in the number of moments. Multimodal uncertainty can be approximated by simulating an ensemble of possible models \cite{pathak_self-supervised_2019}.

Another approach is to drop the belief distribution and maximize Fisher Information \cite{jiang_fisherrf_2024}, which only requires a point \ac{MLE} and is used extensively in optimal experimental design. Recent work uses this approach to refine a simulator from real world data \cite{memmel_asid_2024} and plan informative trajectories in contact-rich environments \cite{sathyanarayan_behavior_2025}. However, maximizing Fisher information alone does not take into account previously observed information over multiple actions. Ergodic exploration \cite{abraham_data-driven_2018, abraham_ergodic_2017} can be used to plan a sequence of actions such that, while more time is spent in areas likely to have high information, the entire state space is eventually explored. \ac{EIG} (see \secref{sec:eig}), derived from Fisher information, explicitly quantifies how much information an action is likely to obtain relative to previously observed information. It has been used in visual model building for active view selection of static objects \cite{jiang_fisherrf_2024, jiang_ag-slam_2024}. This work takes this procedure and implements it for tactile sensing of dynamic objects.

\subsection{Tactile Active Exploration}
Multiple works have tackled the problem of tactile active exploration. The most common application is the modeling of an unknown surface using poking or sliding contact from an optical sensor \cite{hu_active_2024, ketchum_active_2024, yi_active_2016}, as thoroughly collated by a recent survey \cite{schneider_tactile_2025}. It is also common to estimate the geometry of 3d static objects \cite{shahidzadeh_actexplore_2024, driess_active_2017}. When learning both pose and geometry, the latter tends to be treated as a classification problem by maintaining a discrete set of possible object geometries \cite{kim_tactile_2024, xu_tandem3d_2023, zhang_active_2017}. One work simultaneously estimates the pose and geometry of specifically planar objects \cite{suresh_tactile_2021}. To our knowledge, ours is the first work that learns the pose and geometry (from a continuous class of convex shapes) of a 3D rigid body that is free to move during deployment.

\section{Problem Formulation}
\label{sec:problem}
Consider a finite workspace with a rigid known ground plane, unknown convex rigid object, and a robot with known convex end-effectors located at $r_t$, each fitted with a tactile sensor. They report a contact boolean $c_t \in \{0, 1\}$ and a measured contact normal direction $\hat{n}_{t,m} \in S^2$. The goal is to determine the current pose of the object $x_T$ and its geometry $\theta$ in as few actions as possible. In practice, our action space will be a a parameterized trajectory described in more detail in \secref{sec:action}. Given the possible geometric symmetries in the object, the success metric will be the \ac{bCH} between the estimated $\tilde{S}(\tilde{x}_T, \tilde{\theta})$ and ground-truth $S^*(x^*_T, \theta^*)$ surfaces.
\begin{align}
    C_{AB} &= \sum_{a\in A}\min_{b\in B}\twonormsq{a-b}\ ;\ bCH=C_{\tilde{S}S^*}+C_{S^*\tilde{S}}
\end{align}
The interaction protocol is summarized in \figref{fig:block_diagram}.
\begin{enumerate}
    \item \textbf{Learning:} (\secref{sec:vimp}) At current time $T$, given some observed dataset $\Data_{obs} = \bigcup_{t\in[0,T]}\{r_t, c_t, \hat{n}_{t,m}\}$, produce some estimate of the object pose and shape. In practice, this will be the minimum of some loss function $\Loss$.
    \begin{equation}
        \tilde{\theta}, \tilde{x}_T = \arg\min_{\theta,x_T}\Loss(\theta, x_T, \Data_{obs})
    \end{equation}
    \item Evaluate \ac{bCH}
    \item \textbf{Exploration:} (\secref{sec:eig}) Given some known horizon $H > T$, choose a trajectory $r_{T\cdots H}$ and collect new data $\Data_{acq} = \bigcup_{t\in[T,H]}\{r_t, c_t, \hat{n}_{t,m}\}$.
    \item Repeat with $\Data_{obs} \leftarrow \Data_{obs} \cup \Data_{acq}$
\end{enumerate}

\subsection{Sensor Model}

We model a discrete logistic likelihood for $c_t$, where the logit $\phi_t(\theta, x_t, r_t)$ is the signed distance between the robot and the object and $\alpha$ is an uncertainty hyperparameter. For a given value of $c_t$, this likelihood approximates stepwise behavior about $\phi_t = 0$ with a sigmoid logistic function.
\begin{align}
\label{eq:bool}
    p(c_t | \theta, x_t, r_t) = \frac{c_t + (1-c_t)e^{\alpha\phi_t}}{1 + e^{\alpha\phi_t}}
\end{align}

\noindent If $c_t=1$, we model our measured normal $\hat{n}_{t,m}$ with a Gaussian\footnote{While a more appropriate likelihood in $S^2$ may be a univariate Gaussian about the dot product $1 - \hat{n}_{m,t} \cdot \hat{n}_t$ with variance $\sigma^2$, the two are equivalent for diagonal covariance $\Sigma_n=\sigma^2*I$, though the gradients differ in $\R^3$ before normalization. The 2-norm enables the easy calculation of Eq. \ref{eq:obsinfo} that is independent of $\hat{n}_{m,t}$.} likelihood with respect to the expected contact normal $\hat{n}_t(\theta, x_t, r_t)$ with covariance hyperparameter $\Sigma_n$.
\begin{equation}
\label{eq:norm}
\begin{aligned}
    p(\hat{n}_{t,m}|\theta, x_t, r_t) &= \frac{c_t\exp(-0.5\normsq{\hat{n}_{t,m}-\hat{n}_t(\theta, x_t, r_t)}_{\Sigma_n^-1})}{(2\pi)^{3/2}\sqrt{\det\Sigma_n}} \\
    &+ (1 - c_t)\delta(\hat{n}_{t,m})
\end{aligned}
\end{equation}
\noindent While the second term with the Dirac delta function $\delta$ is included here for completeness, it does not involve any object or action parameters and will be omitted going forward.

\section{Learning with a Violation-Implicit Loss}
\label{sec:vimp}
A natural step is to seek to minimize the negative log-likelihood of our sensor model. We take the log of the product of Eq. \ref{eq:bool} and \ref{eq:norm}, bringing the discrete selectors $c_t$ and $(1-c_t)$ out of the logarithm and dropping additive constants.
\begin{equation}
\label{eq:loss}
\begin{aligned}
    \Loss_t = \frac{c_t}{2}\normsq{\hat{n}_{t,m}-\hat{n}_t}_{\Sigma^-1} + \left[(c_t-1)\alpha\phi_t + \log(1 + e^{\alpha\phi_t})\right]
\end{aligned}
\end{equation}
\noindent However, via $\hat n$ and $\phi$, $\Loss_t$ depends on the object state $x_t$, so the sum over the trajectory must be optimized under physical constraints.
\begin{equation}
\label{eq:loss_const}
\begin{aligned}
    \tilde{\theta},\tilde{x}_T = &\arg\min_{\theta, x_T}\sum_t\Loss_t\ s.t.\ x_{t+1} = f_\theta(x_t, r_t, \lambda_t) \\
    &\lambda_t = \arg\min_\lambda g_\theta(x_t, r_t, \lambda)\ s.t.\ \lambda \in \mathcal{FC}
\end{aligned}
\end{equation}
\noindent Where $f_\theta$ are the discrete-time dynamics as a function of contact impulses $\lambda_t$. These can themselves be written as the minimizer of a function $g_\theta$ \cite{anitescu_optimization-based_2006} representing other physical constraints (e.g. non-penetration) discussed in more detail below. Under the assumption of Coulomb friction, impulses are constrained to the friction cone $\mathcal{FC} = \{\twonorm{\lambda_t^f} \leq \mu\lambda_t^n\}$, where the total frictional impulse $\twonorm{\lambda_t^f}$ is bounded by a constant multiple of the normal impulse $\lambda_t^n$, sliding in equality and sticking otherwise.

One could optimize this loss function through shooting and differentiable simulation. With parameter $x_0$, compute $x_t = (f_\theta)^t(x_0)$ and perform gradient descent through $(f_\theta)^t$ to find $\tilde{x}_0$. Then $\tilde{x}_T = (f_\theta)^T(\tilde{x}_0)$. As discussed in related work \cite{bianchini_generalization_2022}, this approach can produce gradients that are near-0 in some regions and near-infinite in others. This is especially true with rigid contact dynamics, longer trajectories, and simulations with a small timestep $\Delta t$.

Intuitively, this procedure is analogous to fitting a Heaviside step function $y=Hs(x-\theta)$ using the square-error loss $\sum_\Data\twonormsq{y_\Data-Hs(x_\Data-\theta)}$, capturing error in the y-direction, which has a gradient of 0 almost everywhere and is discontinuous when data lies at the estimate of the parameter $\theta$. An alternative is to minimize the distance between data and the closest point on the function, or ``graph distance'' $\sum_\Data\min_x\twonormsq{(x_\Data,y_\Data)-(x,Hs(x-\theta))}$. At the cost of a (potentially expensive) inner optimization, the loss gradient is now a finite value (or bounded set) everywhere.

Approximately minimizing graph distance is the essence of the violation-implicit loss, and has been used in geometry learning given a known object trajectory \cite{pfrommer_contactnets_2020}. The analogous procedure in our setting is to bring the near-discontinuous dynamics and physical constraints into the loss function as penalties, and the inner optimization happens over the contact impulses $\lambda$.
\begin{equation}
\label{eq:vimp}
\begin{aligned}
    \Loss_v(\theta, x_t, \Data) = \sum_t\min_{\lambda_t}\ &\Loss_t + \twonormsq{x_{t+1} - f_\theta(x_t, \lambda_t, r_t)} \\
    &+ g_\theta(x_t, \lambda_t, r_t)\ s.t.\ \lambda_t \in \mathcal{FC}
\end{aligned}
\end{equation}
\noindent Without dynamics as a hard constraint, the entire trajectory $x_t$ can be treated as decision variables, reducing the instability of differentiating through a large $T$ at the cost of introducing non-physical local minima in the optimization. In practice, we mitigate the latter using multiple trajectory initializations for the gradient descent. Following \cite{pfrommer_contactnets_2020}, our physical loss terms $g_\theta$ include the following.
\begin{enumerate}
    \item Complementarity: $\lambda^n_t\phi_t$, i.e., contact force only happens when objects touch.
    \item Maximum Power Dissipation: $\twonormsq{v^s_t}\lambda^n_t + (v^s_t)^T\lambda^f_t$, i.e., friction force opposes the contact sliding velocity $v^s_t$ on the boundary of $\mathcal{FC}$
    \item Non-Penetration: $\min(0, \phi_t)^2$
    \item Inelasticity (new in this work): $\lambda^n_t\max(v^n_t, 0)$, i.e., the normal force opposes the normal velocity $v^n_t$ and cannot return energy to the object
\end{enumerate}
\noindent While \cite{pfrommer_contactnets_2020} was agnostic to elasticity given the known trajectory, since we are estimating an unknown trajectory, we found it useful to assume that all of our collisions would be nearly inelastic. Without this term, the object could spontaneously gain energy from the ground contact force, bouncing non-physically. With the above loss terms, our inner optimization loop in $\lambda$ was a \ac{QP} with a second-order cone constraint that could be solved quickly in Cvxpylayers \cite{cvxpylayers2019}. Note that, by the envelope theorem, explicitly differentiating through the $\min_\lambda$ is unnecessary, further speeding up the backwards pass.

\section{Maximizing Expected Information Gain (EIG)}
\label{sec:eig}
\noindent In \secref{sec:prelim_eig}, we first review \ac{EIG} as a popular metric for general active exploration \cite{kirsch_unifying_2022, jiang_fisherrf_2024, jiang_ag-slam_2024}. In \secref{sec:compute_eig}, we detail the specific formulation for our setting: active tactile exploration with a dynamic unknown object and a violation-implicit loss (\secref{sec:vimp}).
\subsection{Preliminaries}
\label{sec:prelim_eig}
As described in \secref{sec:related_ae}, active exploration requires some estimate of the information present in observed data $\Data_{obs}$, some expectation of information that may be available in future data $\Data_{acq}$, and a scalar metric quantifying the utility of that information. A natural step is to consider belief prior $p(\omega | \Data_{obs})$ and posterior $p(\omega | \Data_{obs}, \Data_{acq})$ distributions over model parameters $\omega = \{\theta, x_T\}$. In particular, a common approach is to look at the entropy $H$ of these distributions.
\begin{align}
    H[\omega | \Data] := \E_p[-\log p(\omega | \Data)]
\end{align}
\noindent A chief difficulty is estimating the entropy of the posterior, as the acquired data itself is a random variable sampled from a likelihood distribution $p(\Data_{acq}|\omega, r_t)$.

For a Gaussian, the entropy can be expressed in terms of the log-determinant of the covariance: $H[\mathcal{N}(\mu, \Sigma)] = 1/2 \log \det\Sigma + const$. This is why it is common to use the (co)variance of the belief as a heuristic for high-information regions. But for dynamic objects, propagating a Gaussian prior through time to get that covariance is difficult. The result is fundamentally multi-modal, and just fitting a Gaussian can be completely non-physical. However, it is possible to fit a Gaussian to a second-order approximation of an arbitrary distribution about \emph{any modal peak}. In that case, the effective covariance is equal to the Hessian of the log of the distribution about that peak (\cite{kirsch_unifying_2022}, Prop. 3.5).
\begin{equation}
\label{eq:entropy_approx}
\begin{aligned}
    H[\omega | \Data] &\approx \frac{1}{2}\log\det\nabla^2_\omega(-\log(p(\omega|\Data)))\big|_{\tilde{\omega}} + c
\end{aligned}
\end{equation}

We now have a local approximation of entropy that only requires a local maximum a posteriori (MAP) estimate $\tilde{\omega}$ of the model parameters. Given an uninformative prior (i.e. $Cov(\omega|\emptyset)^{-1}\rightarrow0$), the Hessian of the posterior equals the Hessian of the likelihood, and $\tilde{\omega}$ becomes a local \ac{MLE}.
\begin{equation}
\begin{aligned}
    (\ref{eq:entropy_approx}) &= \frac{1}{2}\log\det\nabla^2_\omega(-\log(p(\Data|\omega)))\big|_{\tilde{\omega}} + c \\
    &:= \frac{1}{2}\log\det\mathcal{I}(\Data, \tilde{\omega}) + c
\end{aligned}
\end{equation}

\noindent The Hessian of the negative log-likelihood about an \ac{MLE} is called \emph{observed information}. Fisher information $\mathcal{F}(\Data, \tilde{\omega}) = \E_{p(\Data)}\mathcal{I}(\Data, \tilde{\omega})$ is defined as the expected value of the observed information from future data.
\ac{EIG}, which is defined to be the difference in entropy between the belief prior and the belief posterior, can be computed without explicitly maintaining either of those distributions by using Fisher and observed information.
\begin{equation}
\label{eq:eig}
\begin{aligned}
    EIG &:= H[\omega|\Data_{obs}] - H[\omega|\Data_{obs}\cup\Data_{acq}] \\
    &\propto \log\det(\mathcal{F}(\Data_{acq})\mathcal{I}(\Data_{obs})^{-1} + \mathbf{I})
\end{aligned}
\end{equation}
\noindent For a full derivation we direct the reader to \cite{kirsch_unifying_2022}, Sec. 3-5.
\subsection{Computing EIG}
\label{sec:compute_eig}
In our setting, the negative log-likelihood is our loss $-\log p(\omega|\Data) = \Loss_t$ (Eq. \ref{eq:loss}), therefore, we can calculate observed information as our loss Hessian at the \ac{MLE} $\tilde{\theta}, \tilde{x}_{0\cdots T}$ computed by our learning scheme.
\begin{align}
    \mathcal{I}(\Data_{obs}) &= \nabla_{\theta, x_T}^2\sum_{t=0}^T\Loss_t(\tilde{\theta}, \tilde{x}_t, r_t)
\end{align}
\noindent Dividing $\Loss_t$ into two components ($\hat{n}_{m,t}$ and $c_t$), the result for each is a term proportional to the Hessian of the logit ($\nabla_\omega^2(\phi_t, \hat{n}_t)$) and a term proportional to the outer product of the gradient of the logit ($\nabla_\omega(\phi_t, \hat{n}_t)\nabla_\omega(\phi_t, \hat{n}_t)^T$). The former can be difficult to calculate. Fortunately, its multiplicand depends on the measurements and is 0 in expectation (noting that $\E[\hat{n}_{m,t}] = \hat{n}_t$ and $\E[c_t] = (1 + e^{\alpha\phi_t})^{-1}$).
\begin{equation}
\begin{aligned}
    &\E_{\hat{n}_{m,t}}[\nabla_\omega^2(\hat{n}_t\cdot(\hat{n}_{m,t}-\hat{n}_t))] = \nabla_\omega^2\hat{n}_t\cdot\cancelto{0}{\E[(\hat{n}_{m,t}-\hat{n}_t))]} \\
    &\nabla_\omega^2(\phi_t)\E_{c_t}[(c_t-1)+e^{\alpha\phi_t}(1+e^{\alpha\phi_t})^{-1}] = 0\ \text{(identically)}
\end{aligned}
\end{equation}
\noindent While observed information is \emph{not} taken in expectation, we omit the above terms, assuming that their contribution will tend towards negligibility as more data is collected.
\begin{equation}
\label{eq:obsinfo}
\begin{aligned}
    \mathcal{I}(\Data_{obs}) \approx \sum_{t=0}^T &\alpha^2\left(\nabla_w\phi_t\left(\frac{e^{\alpha\phi_t}}{(1 + e^{\alpha\phi_t})^2}\right)\nabla_w\phi_t^T\right) \\
    &+ c_t\nabla_\omega\hat{n}_t\Sigma_n^{-1}\nabla_\omega\hat{n}_t^T
\end{aligned}
\end{equation}
\noindent While the gradients $\nabla_\omega$ can be evaluated at the \ac{MLE} trajectory $\tilde{x}_{0\cdots T}$, they must be taken with respect to the parameters we actually care about: $\omega = \{\tilde{\theta}, \tilde{x}_T\}$. Frustratingly, this necessitates quantifying the sensitivity of \emph{past} measurements to the \emph{current} state, as if through a backward simulation $x_{T-1} = f^{-1}(x_T)$. Even with a differentiable simulation, contact dynamics cannot in general be simulated backwards (multiple starting states can lead to the same end state). One could consider taking the dynamics Jacobian $\frac{d x_t}{d x_{t-1}}$ and directly inverting it to try and approximate this sensitivity, but in our experience this leads to severe numerical instability. For this work, we found success computing the sensitivity at each time step with respect to only that time step's state and summing the results, e.g., $\sum_t\nabla_{x_t}\phi_t\nabla_{x_t}\phi_t^T$. This is equivalent to assuming $\frac{d x_t}{d x_{t-1}}=\mathbf{I}$. While sufficient for the sliding object trajectories that emerged in our experiments, we do not believe this approach is likely to work for more chaotic disturbances, such as if the object tumbles over. This is an active avenue for future work.

Fisher information can be similarly calculated. For a given action $r_{T\cdots H}$, we can simulate forward to compute $\tilde{x}_{T\cdots H}$. We note that this future trajectory will be, by construction, a local \ac{MLE}. The results of simulation will perfectly obey physical constraints and simulated measurements are identical to their modeled counterparts, so $\nabla_{x_{T\cdots H}}\Loss_{T\cdots H} = 0$.
\begin{align}
    \mathcal{F}(\Data_{acq}) &= \E_{c_t, \hat{n}_{m,t}}\left[\nabla_\omega^2\sum_{t=T}^H\Loss_t(\tilde{\theta}, \tilde{x}_t, r_t)\right]
\end{align}
\noindent The computation is identical to Eq. \ref{eq:obsinfo}, with the approximation becoming equality as Fisher information is taken in expectation. The same note from the previous paragraph applies for gradient computation with respect to $\tilde{x}_T$. Differentiable calculation of $\phi$ and $\hat{n}$ (itself a \ac{QP}) were implemented with Jaxopt \cite{jaxopt_implicit_diff}. With both Fisher and observed information, we can directly compute \ac{EIG} with Eq. \ref{eq:eig}.

\section{Experiment Setup}
\subsection{Action Space}
\label{sec:action}
As mentioned in \secref{sec:problem}, our action space is a parameterized robot trajectory. For each of our robot's 2 end-effectors, we execute a 1s straight line trajectory that approaches our guess of the object's centroid from a fixed radius. To avoid sensor collisions, each finger is restricted to the +X or -X half-space (bounded from below by the ground). Therefore, our final space is composed of 2, 2D approach angles ($S^2 \otimes S^2$). In practice, we use the ``right-ascension'' (a rotation about +Z from the +/-X-axis) and ``declination'' (a rotation towards the +Z axis). On the real robot, both rotations are limited to $[-\pi/4,\pi/4]$ such that all collisions are likely to happen with the tactile sensor's surface, though in simulation we relax the bounds to $[-2\pi/5,2\pi/5]$.

\ac{EIG} is optimized using an iterative Gaussian cross-entropy method. Initially, a set of $n_{samp}$ actions are sampled uniformly. We the construct a new sampling distribution by sorting the actions by EIG and taking a Gaussian fit of only the best $n_{best}<n_{samp}$. After the last iteration, the final action is chosen uniformly randomly from the final $n_{best}$.

For the computation of \ac{bCH} for evaluation only, FoundationPose \cite{wen_foundationpose_2024} visually computes the ground-truth object pose at 30Hz. As we do not have any size prior for the object, we ignore the initial uniform sweep of the workspace and only begin counting actions at the moment of first contact. In practice, this means randomly selecting an action for one finger centered on the ground truth position.

\begin{figure}[t]
    \centering
    \includegraphics[width=\linewidth]{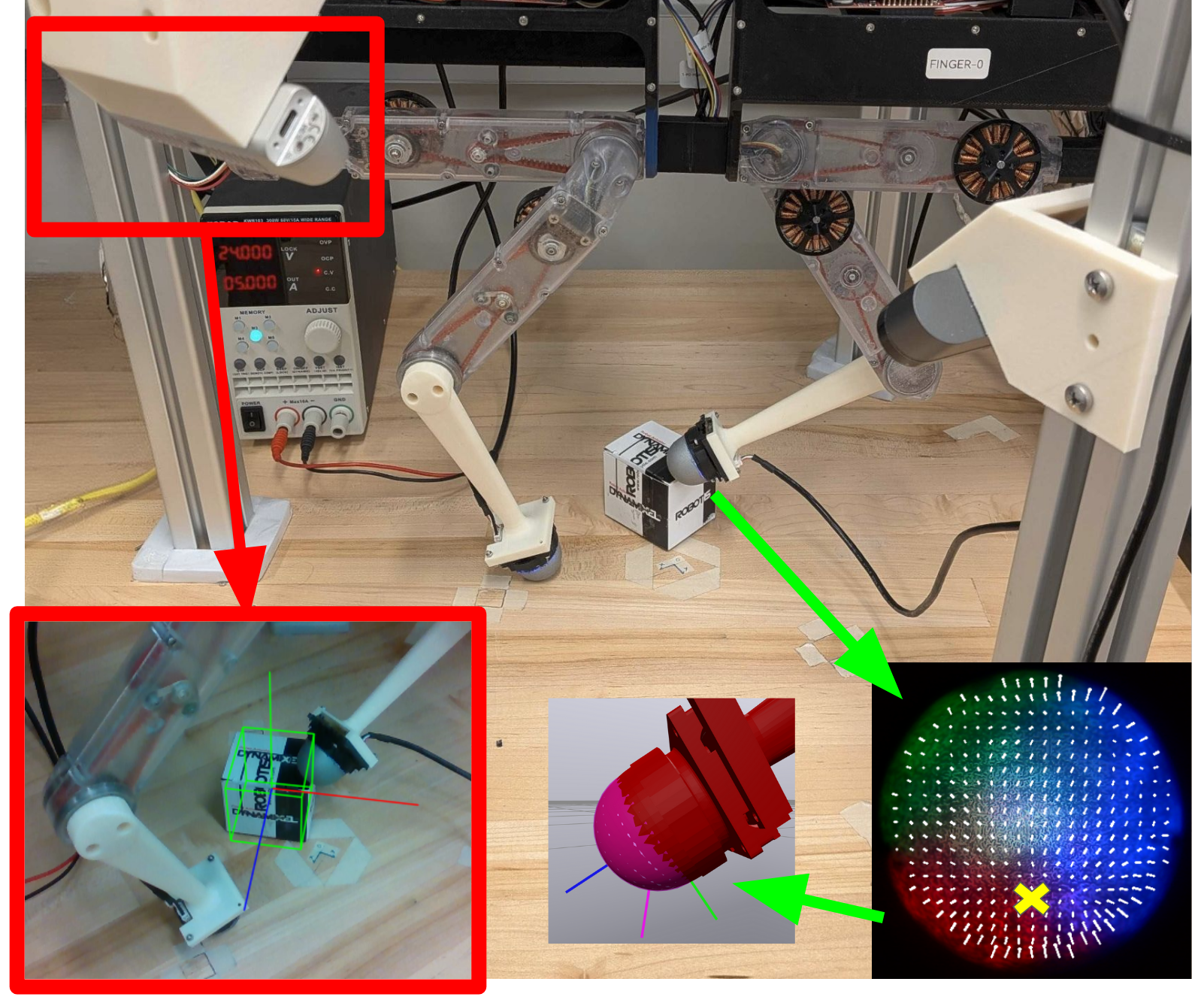}
    \caption{Hardware setup: a Trifinger \cite{wuthrich_trifinger_2021} modified to have 2 fingers directly opposing each other, each instrumented with a Densetact \cite{do_densetact_2023} sensor. Densetact images (green) can be used to locate the center-of-pressure (yellow), which in turn can be projected onto the nominal model to compute the measured normal vector $\hat{n}_{m,t}$ (pink). A single RealSense running FoundationPose (red) is used for evaluation \emph{only} to calculate bCH.}
    \label{fig:setup}
    \vspace{-0.1cm}
\end{figure}

\subsection{Hardware Setup}
Our hardware setup is summarized in \figref{fig:setup}.
We utilize a Trifinger robot \cite{wuthrich_trifinger_2021} modified into a dual-finger configuration 180-degrees apart. Each finger is instrumented with a Densetact \cite{do_densetact_2023}, which streams sensor images at 30Hz. Tactile data acquisition leveraged the correlation between observed fingertip motion primitives from each frame and applied forces (normal, shear, and torsional). Contact motion was computed using the optical flow algorithm \cite{teed_raft_2020} between the current frame and a reference frame with no contact. We perform a Helmholtz-Hodge decomposition \cite{tactile_hhd} of the resulting vector field to separate out the irrotational component. Empirically, we find the surface deformation to be correlated with the scalar potential $\psi(u,v)$ of this field. The center-of-pressure (CoP) can be measured by taking a weighted average $\sum_{u,v}\psi(u,v) \cdot (u,v)$, and the contact normal can be imputed from a model of the sensor's surface. For the contact boolean, we place a threshold on $\psi$ near the CoP.

\subsection{Object Parameterization}

The object pose in $SE(3)$ is parameterized as a quaternion and a displacement. Since the former is normalized before every computation, for the calculation of Fisher and observed information we omit the scalar component.

In this work we consider 2 convex object parameterizations. The cuboid is parameterized in $\R_{>0}^3$ by its side lengths. The convex polytope is parameterized by the convex hull of its vertices in $\R^{3\times n_v}$, where in our experiments $n_v=20$. One benefit of \ac{EIG} is that the object can be re-parameterized at any time. During gradient descent, if any point moves internally, off of the convex hull, it becomes effectively invisible to both the loss function and the final bCH calculation, permanently reducing the expressiveness of the polytope parameterization. To avoid this, we can replace it with a random point on the convex hull.
We also add a regularizer that encourages the polytope to have at least 3 vertices on the ground at the end of the trajectory, as the violation-implicit loss otherwise does not penalize instability.

\section{Results}
\begin{figure}[t]
    \centering
    \includegraphics[width=\linewidth]{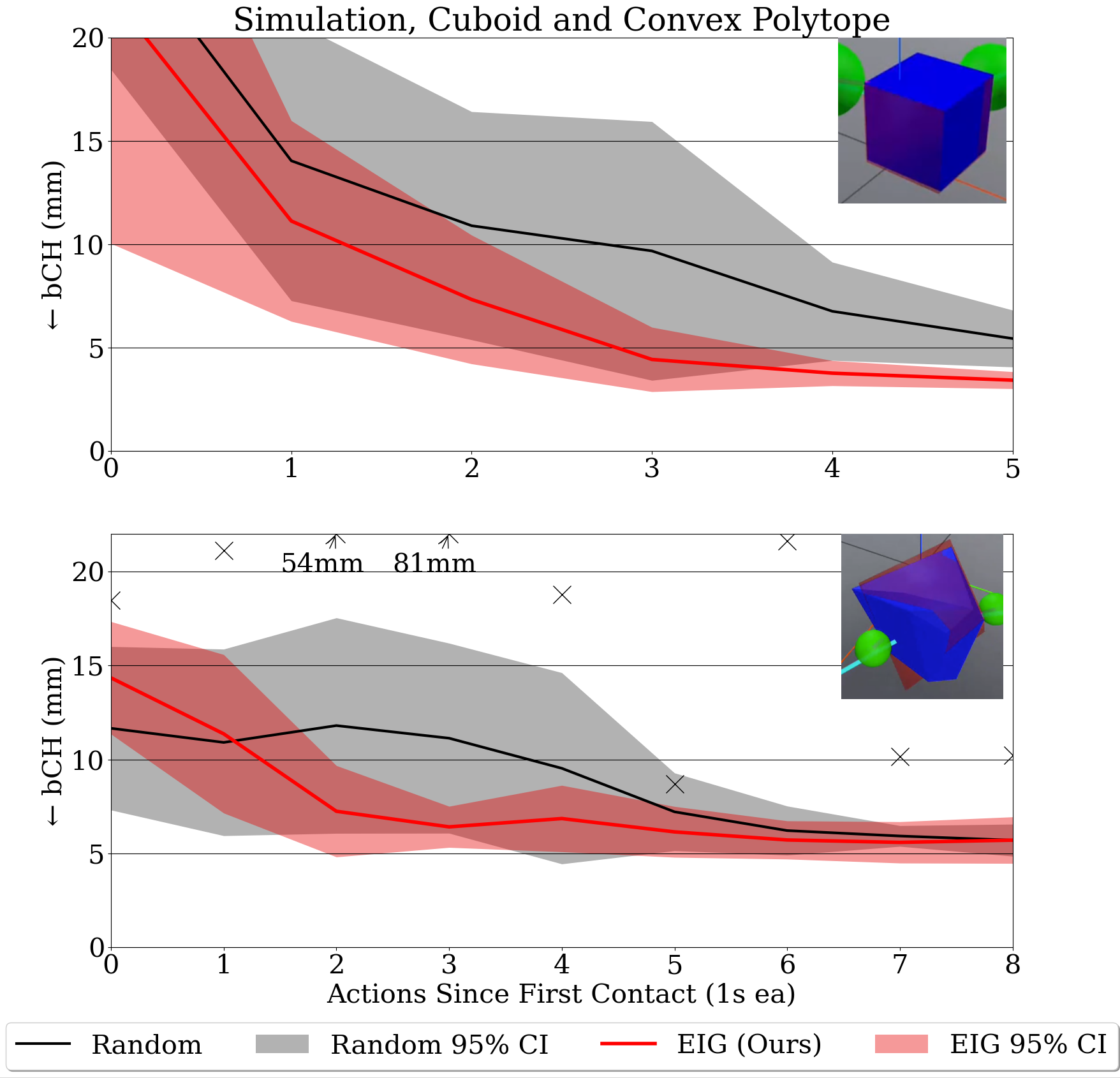}
    \caption{\ac{bCH} between the ground truth and estimated shapes for cuboid (top) and convex polytope (bottom) parameterizations in simulation. Shaded region is the 95\% confidence interval for the mean of a normal distribution ($N=10$). One outlier trial for random (marked with $\times$) does not factor into pictured 95\% CI.}
    \label{fig:sim_res}
    \vspace{-0.1cm}
\end{figure}
\subsection{Simulation Results}
We first evaluated our method in a Drake \cite{tedrake_manipulation} simulation running at 1kHz (with data still reported at 30Hz). For the cuboid parameterization, our ground truth was a box with a randomized shape between 3 and 6cm, with our initial guess randomly sized between 1 and 3cm (in practice, a smaller initial guess lead to more stable learning). For the polytope parameterization, we used Platonic solids of fixed size (2cm bounding radius) as the ground-truth, and our initial guess was always 20 random vertices sampled from the 1cm sphere. Each trial, we initialized both the ground truth and our initial guess to a random pose within 2cm of the center of the workspace. After the initial random action that made contact with the ground truth, we took 5 (for cuboid) or 8 (for polytope) subsequent actions, recording bCH after each one. The results are summarized in \figref{fig:sim_res}.

For the cuboid parameterization, \figref{fig:learning} (top) shows an example 5-action learning trajectory. Observed information tends to jump quickly, as with only 3 geometry parameters a single action with low-variance data hitting the corners is sufficient to learn all of them with high certainty. While uniform random exploration does reach $5mm$ within 5 actions (5s of data), \ac{EIG} exploration consistently reaches $<4mm$ in the same time. At time step 5 this represents a significant improvement (two-tailed t-test with independent samples, $p < 0.05$ with a Bonferroni correction of 5 for all time steps, $N=10$).

For the polytope parameterization, \figref{fig:learning} (middle) shows 5 actions (excluding actions 3, 6, and 7 for brevity) of an example 8-action learning trajectory. Both EIG and Random exploration reach a bCH of 5mm consistently within 8 actions (8s of data). While not rising to the same statistical significance, a clear difference can still be seen between \ac{EIG} and Random, with the latter requiring 3 extra actions for the 95\% confidence interval to drop below 1cm. Note that this interval for Random already omits a catastrophic outlier, where the centroid of the guess was completely outside of the ground truth object, leading to 2 actions without contact before they re-aligned. This is an example of a suboptimal local minimum arising from the learning scheme. Combined with an observable reduction in the variance of bCH across all actions, we can confidently say that the data supports the claim that \ac{EIG}-maximizing exploration provides a more consistent learning trajectory. At the same time, even Random exploration still achieved consistent results with 5-8s of tactile data. 

\subsection{Real Robot Results}
\begin{figure}[t]
    \centering
    \includegraphics[width=\linewidth]{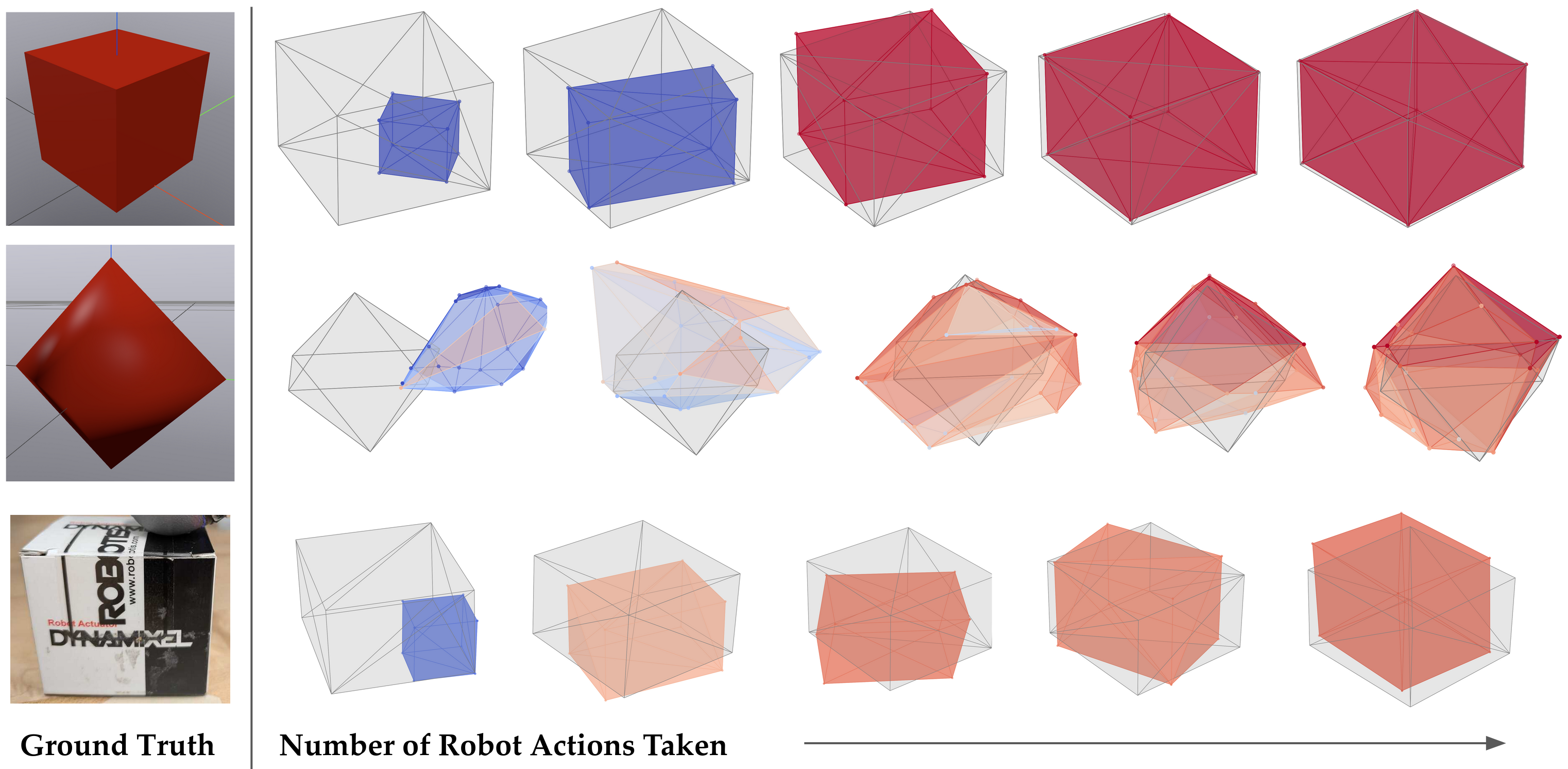}
    \caption{Example learning trajectories for (from top to bottom): a simulated cuboid ground truth with a cuboid parameterization, a simulated regular octahedron ground truth with a 20-vertex polytope parameterization, and a real cuboid ground truth with a cuboid parameterization. Colors represent per-vertex (for the polytope) or total observed information $\log\det\mathcal{I}(\Data_{obs})$. Each image is re-centered on the ground truth, which moved 1-2cm over the course of the trial.}
    \label{fig:learning}
    \vspace{-0.1cm}
\end{figure}

On our real Trifinger and Densetact setup, we used a cuboid ground truth and parameterization. The ground truth object had dimensions $5.8 \times4.9 \times5.3$cm and weighing $\sim0.5$kg. The object generally moved $1-2$cm from its starting pose during each trial. Initialization was randomized as in simulation, with 5 actions performed after first contact. \figref{fig:learning}(bottom) shows an example 5-action learning trajectory.

\begin{figure}[t]
    \centering
    \includegraphics[width=\linewidth]{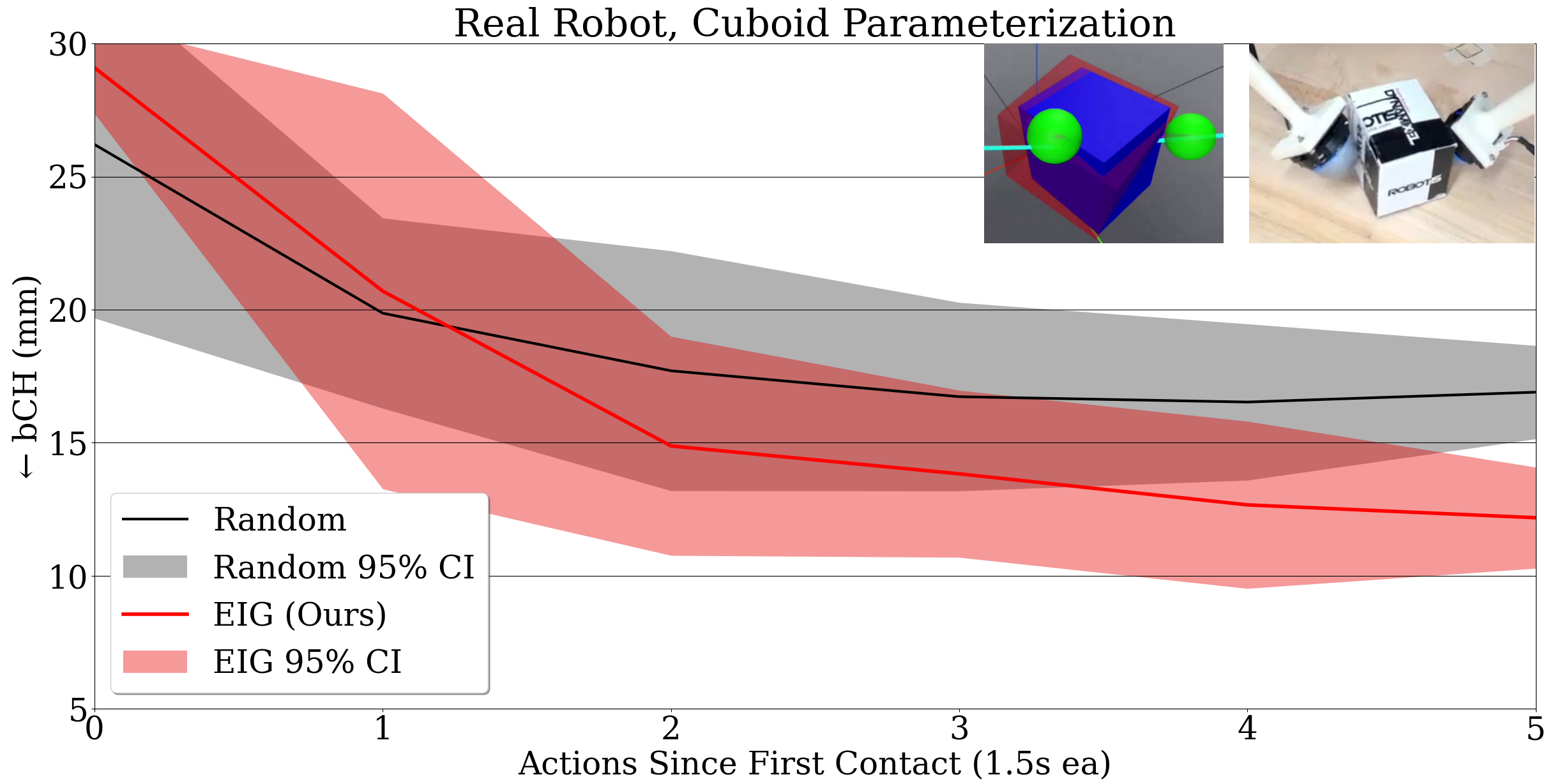}
    \caption{\ac{bCH} between the ground truth and estimated shapes on the real robot. Shaded region is the 95\% confidence interval for the mean of a normal distribution ($N=6$).}
    \label{fig:real_res}
\end{figure}

The results are summarized in \figref{fig:real_res}. While the overall absolute bCH is worse than simulation, a large portion can be attributed to 3-5mm of proprioceptive error in each end-effector. We do still see a significant improvement in \ac{EIG} over Random exploration at time step 5 (two-tailed t-test with independent samples, $p < 0.05$ with a Bonferroni correction of 5 for all time steps, $N=6$). 


\section{Discussion}
\label{sec:discussion}
We have demonstrated that we can identify the geometry and pose of a continuous class of unknown dynamic convex rigid bodies with only 5-8s of tactile contact data. Furthermore, we have shown that our active exploration procedure, by maximizing expected information gain, can lead to significantly faster and more consistent performance.

This procedure is not without limitations. Further hardware improvements can increase performance, e.g., the Densetact struggled to detect light touches, thus limiting our experiments to moderately heavy objects. More generally, \ac{EIG} as a local metric can perform poorly if the \ac{MLE} is a non-optimal local minimum. Our action space, which relied on the estimated object pose, was similarly vulnerable. This required multiple initializations to mitigate, which on top of the increasing dataset size over the course of the experiment slowed down training.
Also, as discussed in \secref{sec:compute_eig}, computing \ac{EIG} with respect to $x_T$ requires quantifying the sensitivity of past measurements to future states, which is unavoidably poorly posed for stiff contact dynamics. Subject to Coulomb friction, no amount of smoothing can change the fact that, if the object is at rest $(z_T=0)$, it could have always been there or just arrived. While an identity dynamics Jacobian was sufficient for this work, further investigation into this issue is necessary.

With that said, this success can still be expanded to more fully demonstrate the model-building capability of tactile sensing. For example, while the current approach requires a convex object and end-effector to define a unique collision point, it should be possible to model non-convex objects as a union of convex shapes, further extending the space of possible geometry representations. Additionally, this work can be combined with a vision-based model-building solution \cite{jiang_fisherrf_2024, bianchini_vysics_2025} to create a robust multi-modal system. This is a step towards the goal of a robot that can use all of the sensors at its disposal to model previously-unseen arbitrary objects in its environment throughout its entire deployment lifetime.


\balance
\section*{Acknowledgments}
\ifthenelse{\boolean{anonymized}}{
Acknowledgments Omitted for Anonymous Review
}{
This work was supported by an NSF CAREER Award under Grant No. FRR-2238480 and the RAI Institute. The authors extend their gratitude to Hrishikesh Sathyanarayan
and Ian Abraham from the CoCuRo Lab at the University of Sydney and Yale University for all the insightful discussions early in the conception of this work.
}

\bibliographystyle{IEEEtran}
\bibliography{IEEEabrv, references}

\end{document}